\documentclass[10pt,twocolumn,letterpaper]{article}

\usepackage{cvpr}              

\usepackage{graphicx}
\usepackage{amsmath}
\usepackage{amssymb}
\usepackage{booktabs}
\usepackage{tabularx}
\newcommand{\darr}{$\downarrow$}
\newcommand{\uparr}{$\uparrow$}

\usepackage{tikz}

\usepackage[pagebackref,breaklinks,colorlinks]{hyperref}

\usepackage[capitalize]{cleveref}
\crefname{section}{Sec.}{Secs.}
\Crefname{section}{Section}{Sections}
\Crefname{table}{Table}{Tables}
\crefname{table}{Tab.}{Tabs.}

\def\cvprPaperID{32} 
\def\confName{CVPR}
\def\confYear{2025}

\begin{document}

\title{Efficient 3D Full-Body Motion Generation from Sparse Tracking Inputs with Temporal Windows}

\author{Georgios Fotios Angelis$^{1}$\footnotemark[1] \quad Savas Ozkan$^{2}$\thanks{Equal Contributions. \texttt{savas.ozkan@samsung.com}} \quad Sinan Mutlu$^{2}$\footnotemark[1] \quad Paul Wisbey$^{2}$ \quad Anastasios Drosou$^{1}$ \\  \quad Mete Ozay$^{2}$ \\
$^{1}$CERTH \quad $^{2}$Samsung R\&D Institute UK \\
}

\maketitle

\begin{abstract}

To have a seamless user experience on immersive AR/VR applications, the importance of efficient and effective Neural Network (NN) models is undeniable, since missing body parts that cannot be captured by limited sensors should be generated using these models for a complete 3D full-body reconstruction in virtual environment. However, the state-of-the-art NN-models are typically computational expensive and they leverage longer sequences of sparse tracking inputs to generate full-body movements by capturing temporal context. Inevitably, longer sequences increase the computation overhead and introduce noise in longer temporal dependencies that adversely affect the generation performance. In this paper, we propose a novel Multi-Layer Perceptron (MLP)-based method that enhances the overall performance while balancing the computational cost and memory overhead for efficient 3D full-body generation. Precisely, we introduce a NN-mechanism that divides the longer sequence of inputs into smaller temporal windows. Later, the current motion is merged with the information from these windows through latent representations to utilize the past context for the generation. Our experiments demonstrate that generation accuracy of our method with this NN-mechanism is significantly improved compared to the state-of-the-art methods while greatly reducing computational costs and memory overhead, making our method suitable for resource-constrained devices.

   
\end{abstract}

\section{Introduction}
\label{sec:intro}

3D human full-body tracking has great potential across multiple applications, such as sports analytics and interactive gaming. This ability becomes particularly critical in immersive technologies like Virtual Reality (VR) and Augmented Reality (AR), where instantaneous motion capture forms the foundation for true digital interactions. In particular, by translating users' body kinematics into virtual space, these systems can obtain high perceptual realism, allowing users to interact with the virtual elements through movement patterns. This synchronization between real-motion and feedback through sensors fundamentally enhances user embodiment within synthetic environments.

\begin{figure}[!t]
\centering
\begin{tabular}{c}
 
\includegraphics[width=0.45\textwidth]{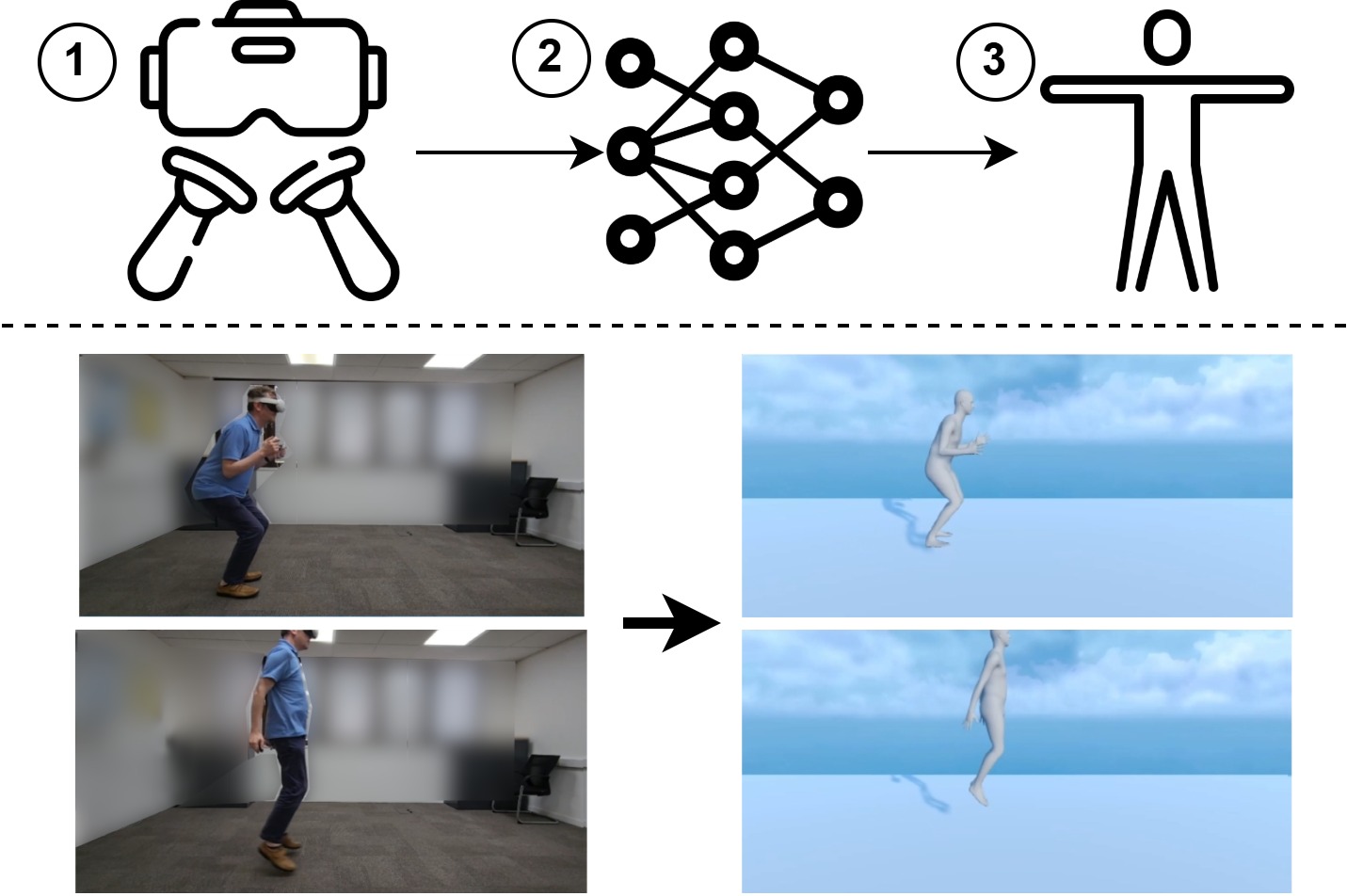}

\end{tabular}
\vspace{-0.8em}
\caption{Visualization of 3D full-body motion generation process using sparse tracking inputs. First, sparse tracking inputs are captured with a VR headset and hand controllers. These inputs are then used with a NN-model to generate 3D full-body motions. In the figure, we also provide real-visual samples from a setup using the Meta Quest-3 headset and our model to illustrate this process.}
\vspace{-1.5em}
\label{fig:visual}
\end{figure}

Hence, it is essential to track the movement of their full-body in high accuracy to fully immerse users in Virtual Reality (VR) and Augmented Reality (AR) systems. However, current solutions are limited by their hardware, which usually consists of a headset and handheld controllers that have sensors to track only certain movements. This means that the movements of other body parts (i.e., lower-body parts) cannot be captured, resulting in an incomplete translation of the full-body into the virtual environment. To this end, this creates a sense of disconnect and makes the virtual world less immersive.

To address the limitations of current VR/AR systems, researchers have investigated various methods to achieve more comprehensive full-body tracking/estimation. Some promising approaches include the use of markerless motion capture systems, which utilize computer vision algorithms to track full-body movements without the need for physical markers or sensors. Other approaches involve the integration of additional wearable sensors which can track other body parts. Recent studies~\cite{von2017sparse} have explored the use of multiple inertial measurement units (IMUs) attached to different parts of a body, such as the wrists, legs, back and head. Similarly, other works~\cite{yang2021real} use only four sensors for the pelvis, head, and hands. Although these approaches are promising, they require specialized setups that must be tailored to specific scenarios, which can be impractical and limiting.

Instead of relying on diverse sensor data from different body parts, an alternative approach is to estimate lower body motion using limited sensor data collected only from the upper body. Advances in deep learning and computer vision have made it possible to accurately estimate full-body motion, leading to the development of more accurate and effective solutions to infer full-body movements. In many extended reality (XR) applications, the number of available sensor signals is often limited to just three: the head and two hands. To address this challenge, researchers have developed various methods, such as AvatarPoser~\cite{jiang2022avatarposer}, which uses a transformer-based architecture to estimate full-body motion from these three tracking inputs. Alternatively, other methods address the problem as a 3D generation task, employing techniques such as flow-based architectures~\cite{Aliakbarian_2022_CVPR} or Variational Autoencoders (VAEs)~\cite{Dittadi_2021_ICCV} to estimate full-body motion from sparse input data.

More recent works have explored the use of diffusion-based architectures for full-body motion estimation. For instance, researchers in~\cite{ dong2024realistic} have proposed novel approaches using diffusion-based models. Another approach~\cite{feng2024stratified} employs separate Vector Quantized Variational Autoencoder (VQ-VAE) codebooks for the lower and upper body parts, leveraging a two-stage diffusion process. The first diffusion model learns VQ-VAE codebooks from sparse sensor data, while the second model generates codebooks for the lower body using the sparse and upper body codebooks created in the first stage. These codebooks are then fed into a decoder model to generate full-body motions. A U-Net shaped architecture is also utilized in~\cite{dong2024realistic} by incorporating Motion Mamba-Blocks.

The success of deep learning models is recognized due to the fact that they can learn nearly all possible solutions from large datasets and their response can be robust to handle complex cases. However, they require significant computational power, which can lead to increased energy consumption, heat, and latency that can adversely affect user experience. An ideal solution should balance high accuracy with efficiency, scalability, and practicality, making it suitable for real-world applications on resource-constrained devices.

For this purpose, an Multi-Layer Perceptron (MLP) network is proposed in~\cite{du2023avatars} which effectively generates 3D full-body motion with a relatively simple architecture compared to other methods. In particular, this method is more suitable for real-time applications because of its moderate memory overhead and computational costs. However, the model still requires longer motion sequences to compute long-range dependencies and contextual relationships between sparse tracking inputs, which results in higher computational costs during the input projection step. Furthermore, this can introduce noisy relationships (i.e., those with higher uncertainty) from past time instances during the generation step, affecting the smoothness of the generated motions. As a result, integrating long sparse input sequences into the model requires a different arrangement to mitigate this issue.


In this work, we present a novel MLP-based method that is suitable for real-time 3D full-body motion generation by balancing the trade-off between accuracy and complexity. In particular, the main objective of our method is to divide the longer sequence of inputs into multiple temporal windows and combines the information from the current window with past windows through the latent representations. This adaptation provides two critical benefits for enhanced motion generation accuracy and faster inference: 

\begin{enumerate}
\item By partitioning long sequences into smaller temporal windows, we compute more focused information extracted from each window, which inherently captures distinct aspects of body kinematics. This enables us to effectively condition past information onto the current motion state, resulting in robust and stable representations (i.e., reduced variance and noise). Moreover, by integrating past information through latent representations rather than feeding it in as raw input data, we can leverage a more refined and abstracted form of past context.

\item Since the projection step is executed on smaller windows of sparse tracking signals, the computation is greatly reduced, particularly in the projection step. Moreover, we observe that temporal windowing of a sequence improves the accuracy of generation by minimizing noise and variance that allows the use of shorter motion sequences for high accuracy, unlike the original model which relies on longer sequences~\cite{du2023avatars}.  

\begin{figure*}[t]
    \centering
    \includegraphics[width=0.7\textwidth]{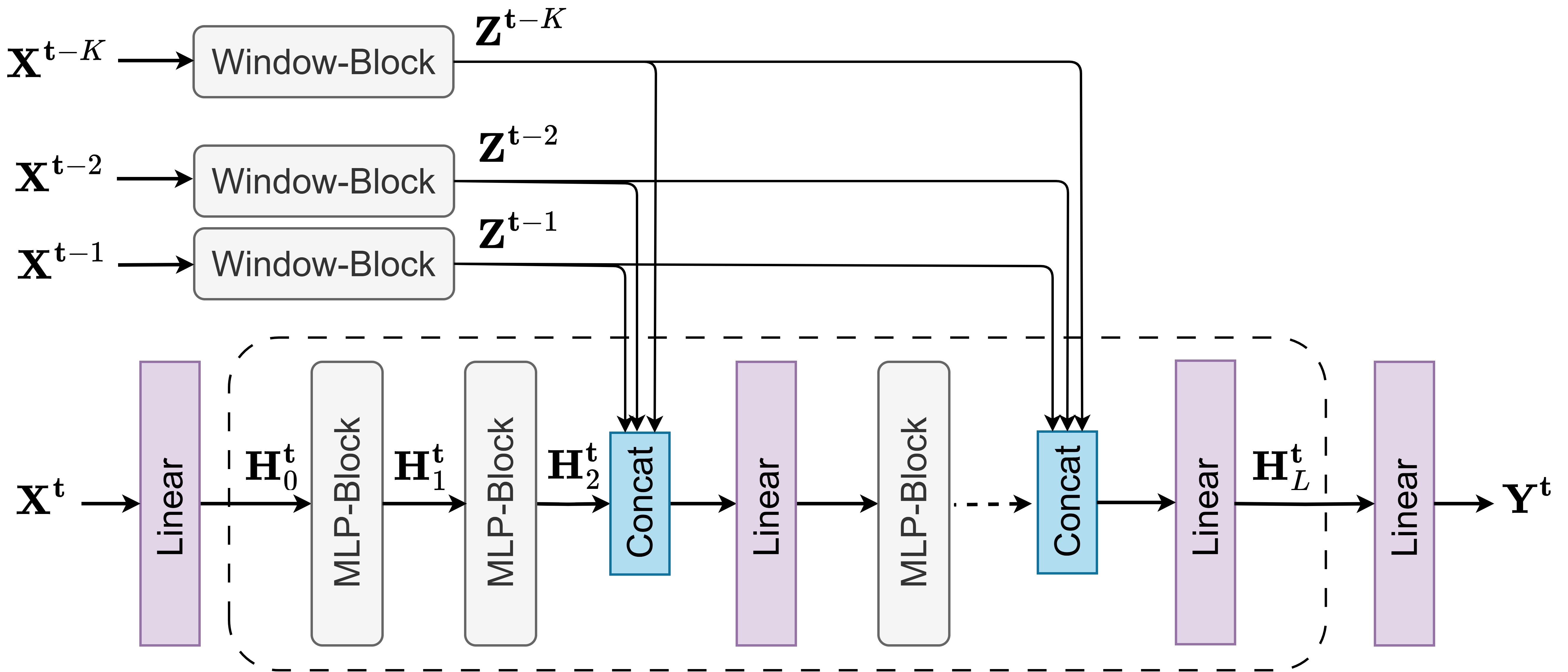}
    \caption{The architecture of our method. The details are  described in Section~\ref{sec:method}.}
    \vspace{-0.8em}
    \label{fig:arch}
\end{figure*}

\end{enumerate}

Ultimately, our method provides both high accuracy and real-time capability, making it an ideal solution for applications requiring fast and reliable performance for head-mounted devices such as the Meta Quest-3 Headset, as illustrated in Fig.~\ref{fig:visual}. Our key contributions are summarized as follows:

\begin{itemize}
\item We propose a novel NN-based mechanism for 3D full-body motion generation, which enhances the accuracy by dividing the long sequence of sparse tracking inputs into smaller temporal windows and merging these past windows into the current motion through latent representations. This mechanism significantly improves the 3D human motion generation accuracy while reducing the Jitter metric from $13.01$ to $7.99$. Details of the mechanism are explained in Sec.~\ref{sec:window}.

\item Our method obtains more efficient results in terms of computational costs and memory overhead compared to the state-of-the-art method \cite{du2023avatars} that uses the same MLP-model, where the overall FLOPs reduce from $0.88$ to $0.19$. Hence, our method can be deployed into head-mounted devices, meeting the minimum requirement on runtime efficiency. This enables a real-time and seamless user experience for immersive interactions, due to its efficiency and high-accuracy.  

\end{itemize}

\section{Proposed Method}
\label{sec:method}

In this section, we first provide the preliminaries of the task for 3D full-body motion generation using sparse tracking inputs. Later, we explain the details of our method that leverages temporal windowing for long sequences to produce reliable 3D full-body motion generation by integrating past data.

\subsection{Preliminaries}


\subsubsection{Problem Definition:} 

The objective of 3D full-body motion generation is to estimate full-body motion $\mathbf{y}^t \in \mathbb{R}^m$ from sparse tracking input $\mathbf{x}^t \in \mathbb{R}^d$ at a time instance $t$ using a model $f$ that is represented by $\mathbf{y}^t=f(\mathbf{x}^t)$. Here, $m$ and $d$ denote the dimensions of full-body motion and sparse tracking input at the time instance $t$, respectively. In this problem, number of sensor outputs is limited to three that are attached to head, left and right hand joint locations. Therefore, sparse tracking input $\mathbf{x}^t$ consists of these signals (i.e., sensor outputs) as $\mathbf{x}=[\mathbf{x}_{\text{head}}^t, \mathbf{x}_{\text{left}}^t, \mathbf{x}_{\text{right}}^t]$, where $[.,.]$ defines the concatenation operation. Furthermore, each signal from different joint location $\mathbf{x}_j^t$ (where $j \in \{\text{head}, \text{left}, \text{right}\}$) is represented by an axis-angle representation and a 3D coordinate representation $\mathbf{p}_j^t$. For the higher accuracy~\cite{jiang2022avatarposer}, the axis-angle representation is first converted to a rotation matrix $\mathbf{R} \in \mathbb{R}^{3\times3}$ and the first-two rows of this matrix are then used to obtain a 6D rotation representation $\theta_i^t$. Also, a 3D linear velocity representation $\mathbf{v}_j^t=\mathbf{p}_j^t-\mathbf{p}_j^{t-1}$ and a 6D rotation velocity representation $\mathbf{w}_j$ (similarly, the first-two rows of $\text{inv}(\mathbf{R}^{t-1})\mathbf{R}^t$ are used, where $\text{inv(.)}$ denotes the matrix inversion operation.) are calculated for temporal smoothness~\cite{jiang2022avatarposer}. To this end, each sparse tracking input is represented by $\mathbf{x}_j=[\mathbf{p}_j, \theta_j, \mathbf{v}_j, \mathbf{w}_j]$. The dimension of sparse tracking input $d$ is $54$ at the time instance $t$.

For the full-body motion $\mathbf{y}^t$, the SMPL body model~\cite{loper2023smpl} is exploited and first $22$ joints are selected as the representation which does not include the hand joints. Specifically, the full-body motion is denoted as $\mathbf{y}^t=[\theta_j^t]_{j=1}^{22}$ that consists of 6D rotation representations $\theta_j^t$. To this end, this results in a total dimensionality $m$ of 132 for the full-body motion representation at the time instance $t$.

\subsubsection{Sequence of Sparse Tracking Inputs:} 
Generating true full-body motion $\mathbf{y}^t$ using only sparse tracking input $\mathbf{x}^t$ at a time instance $t$ is impractical, since the models cannot accurately predict the full-body motion from small portion of lower-body data. To overcome this, previous studies \cite{Feng_2024_CVPR, dong2024realistic} have utilized the sequence of sparse tracking inputs to extract high-level context information that intuitively captures various aspects of body kinematics. Hence, the model $f$ is reformulated as $\mathbf{Y}^\mathbf{t}=f(\mathbf{X}^\mathbf{t})$ to handle the sequence of sparse tracking inputs $\mathbf{X}^\mathbf{t}=[\mathbf{x}^i]_{i=t-T}^t$ and generate the sequence of full-body motions $\mathbf{Y}^\mathbf{t}=[\mathbf{y}^i]_{i=t-T}^t$. Here, $\mathbf{t}$ denotes the temporal window whose length is $T$, covering data from time instance $t$ to time instance $t-T$. If we generalize this window timing concept, $\mathbf{t}-k$ corresponds to the temporal window whose time instances vary from $t-T(k-1)$ to $t-Tk$. To this end, the model generates the sequence of full-body motions using the sequence of sparse tracking inputs within the temporal window $\mathbf{t}$ by learning and inferring high-level contextual information.

\subsubsection{Base MLP-Model:} 
\label{sec:mlp}

\begin{figure}[t]
    \centering
    \includegraphics[width=0.45\textwidth]{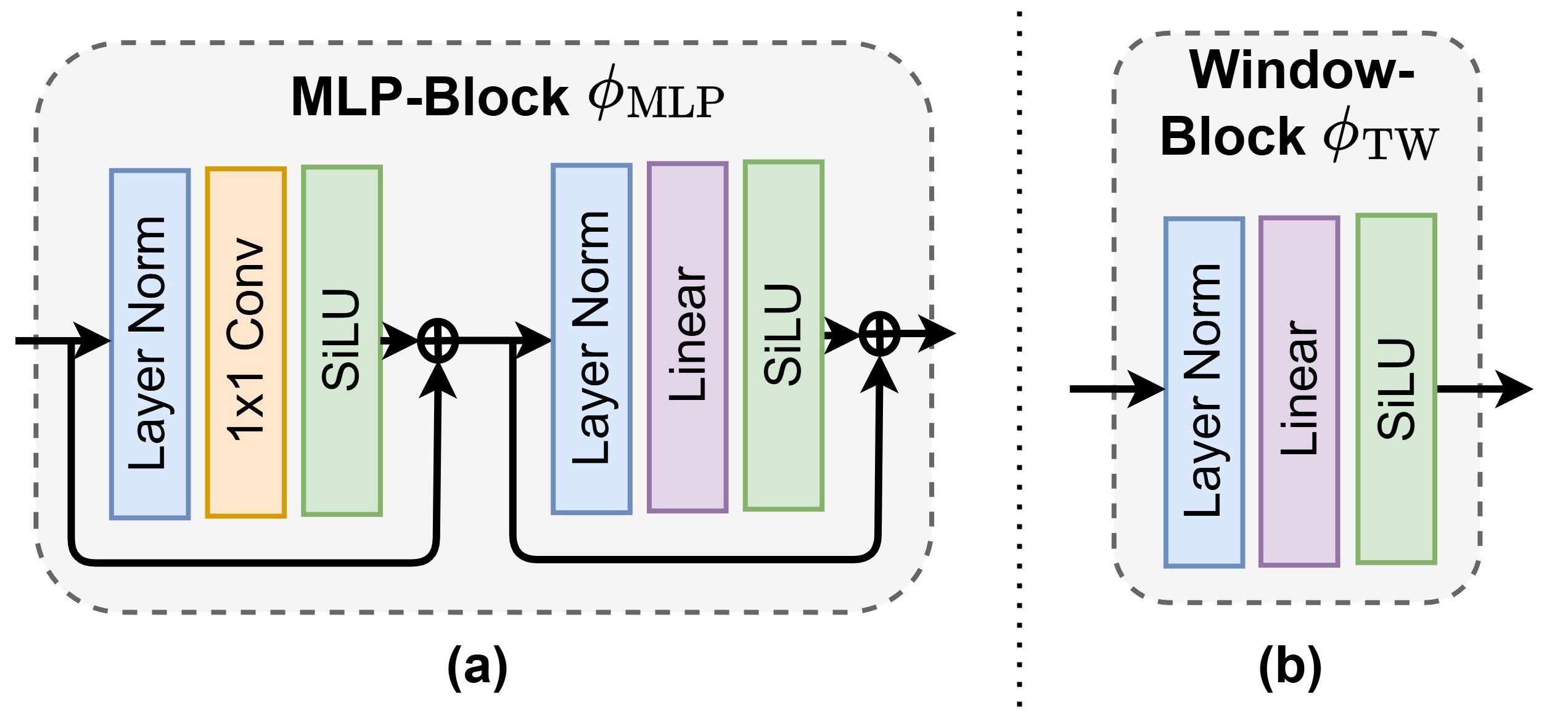}
    \caption{The architecture of (a) MLP-Block (Sec.~\ref{sec:mlp}) and (b) Window-Block (Sec.~\ref{sec:window}).}
    \label{fig:blocks}
    \vspace{-0.8em}
\end{figure}

For 3D full-body motion generation, the objective of NN-models is to achieve high accuracy while minimizing computational overhead, enabling their deployment into mobile devices. MLP-based models are well-suited for this task, as they balance the trade-off between accuracy and complexity. One of the most prominent models for this problem that is based on MLP is proposed in~\cite{du2023avatars}, which comprises four neural network (NN) components: fully connected layers (FC), SiLU activation layers, 1D convolution layers (1D-Conv), and layer normalization (LN). The architecture of MLP-block is illustrated in Fig.~\ref{fig:blocks}. The overall model is composed of $L$ MLP-blocks, each configured using these components. In particular, the 1D convolution and fully connected layers utilize together to extract temporal and spatial information from the sequence of sparse tracking inputs, while skip connections enhance the richness of the representations with incremental representation learning. 

In practice, the sequence of sparse tracking inputs $\mathbf{X}^\mathbf{t}$ is first projected onto a latent representation $\mathbf{H}_0^\mathbf{t}$ using a linear projection matrix $W_p$ by $\mathbf{H}_0^\mathbf{t}=W_p\mathbf{X}^\mathbf{t}$. Later, $\mathbf{H}_0^\mathbf{t}$ is iteratively processed by $L$ MLP-Blocks $\phi_l$ by outputting latent representations $\mathbf{H}_{l+1}^\mathbf{t}=\phi^l_{\text{MLP}}(\mathbf{H}_l^\mathbf{t})$ for all layers. In the final layer, another linear projection matrix $W_f$ is utilized to estimate the sequence of 3D full-body motion predictions $\mathbf{Y}^\mathbf{t}$ using the final-layer latent representation by $\mathbf{Y}^\mathbf{t}=W_f\mathbf{H}_L^\mathbf{t}$. 

\subsubsection{Loss Functions}

Current research on 3D full-body motion generation primarily addresses it as a single-task learning problem, focusing on estimating the $6D$ rotations $\theta_j^t$ for each joint $j$ at a time instance $t$~\cite{du2023avatars, feng2024stratified, dong2024realistic, jiang2022avatarposer, Dittadi_2021_ICCV, zheng2023realistic}. These methods employ a multi-loss function framework, where all losses are centered around minimizing rotation errors. Specifically, a rotation loss $\mathcal{L}_{\theta}$ is used to reduce the difference between the ground-truth rotations $\theta^{t}_{j}$ and the predicted rotations $\hat{\theta^{t}_{j}}$ for all full-body joints, with the goal of improving the accuracy of rotation predictions: 
\begin{equation}
\begin{split}
\mathcal{L}_{\theta} = \frac{1}{T}  \sum_{t=1}^{T} \sum_{j=1}^{22} \| \theta^{t}_{j} - \hat{\theta}_{j}^t \|_1,
\end{split}
\end{equation}

\noindent
where $T$ indicates the length of temporal window and $22$ corresponds to the number of joints in the full-body model. To ensure smooth translations in temporal motions, a rotation velocity loss can also be introduced $\mathcal{L}_{rv}$, which provides an alignment between the predicted rotation velocity and the ground truth. This loss function is designed to encourage smooth and accurate motion generation and is defined as:
\begin{equation}
\begin{split}
\mathcal{L}_{rv} = \frac{1}{T-1} \sum_{t=1}^{T-1} \sum_{j=1}^{22} \| (\theta^{t+1}_j - \theta^t_j) - (\hat{\theta}^{t+1}_j - \hat{\theta}^t_j) \|_1.
\end{split}
\end{equation}

The overall loss function $\mathcal{L}$ used to train NN-models is formulated by
\begin{equation} 
    \begin{split} 
        \mathcal{L} = & \ \lambda_{\theta} \mathcal{L}_{\theta} 
        + \lambda_{rv} \mathcal{L}_{rv} + 0.0001 \mathcal{L}_{reg}
    \end{split} 
    \label{eq:loss}
\end{equation}

\noindent
where $\mathcal{L}_{reg}$ represents the regularization term for L2 weight decay. Furthermore, $\lambda_{\theta}$ and $\lambda_{rv}$ are the coefficients that control the contribution of each loss term to the overall loss function. 

\subsection{MLP-Model with Temporal Windowing}
\label{sec:window}

The limitation of base MLP-model~\cite{du2023avatars} is that it depends on a long sequence of sparse tracking inputs to truly generate full-body motions. This introduces additional computational overhead and noise in temporal dependencies which eventually reduce the performance of the base model. For this purpose, we propose a novel mechanism that consists of NN-blocks, dividing the sequence of sparse tracking inputs into temporal windows and integrating them into the current motion generation process through latent representations. To do so, our mechanism enhances the accuracy of generation and provides faster inference speed. The overall architecture of our method with the temporal windowing mechanism is illustrated in Fig.~\ref{fig:arch}. 

In our method, we employ the base MLP-model~\cite{du2023avatars} to have an efficient and effective full-body motion generation. We also introduce temporal window blocks that process the sequences of sparse tracking inputs from past windows, producing latent representations. These latent representations from past and current windows are then combined using a concatenation operation, allowing to combine information for more accurate and efficient motion generation.

Formally, the window-block $\phi^{k}_{TW}$ for the $k^{\text{th}}$ temporal window takes 
the sequence of sparse tracking inputs $\mathbf{X}^{\mathbf{t}-k}$ from $\textbf{t}-k$ window as input, where $k=1,...,K$. Then, it outputs a latent representation for the $k^{\text{th}}$ temporal window as $\mathbf{Z}^{\mathbf{t}-k}=\phi^{k}_{TW}(\mathbf{X}^{\mathbf{t}-k})$. As noted, for each window-block, a separate MLP-model is utilized in our architecture. The architecture of each NN-block consists of linear layer, layer normalization and SiLU activation layer as illustrated in Fig.~\ref{fig:blocks}. Later, the latent representations from the base MLP-model $\mathbf{H}_l^\mathbf{t}$ and all window-blocks $\mathbf{Z}^{\mathbf{t}-k}, \forall k$ are concatenated at the $l^{\text{th}}$ layer and a linear projection matrix $W_l$ is utilized to obtain a more compact representation before feeding it to the next MLP-block. Here, the window length $T$ and the number of temporal windows $K$ are critical parameters in our method for 3D full-body motion generation. We will investigate these parameters in our experiments to understand how they influence our results.

To this end, our novel method effectively combines current and past sequential information through latent representations by achieving stable, efficient and precise 3D full-body motion generation.

\subsubsection{Optimization}

To optimize the trainable parameters in our method, we leverage the loss function presented in Eq.~\ref{eq:loss}. Differently, we formulate the training step as a multi-task learning problem, instead of using fixed weights for these terms. Specifically, we employ an uncertainty-based technique~\cite{kendall2018multi} to dynamically adjust the weights during training, which helps to reduce the sensitivity of the technique to the weight selection. This allows the technique to learn optimal weights for each loss term, leading to more robust and effective training results.

\begin{table*}[!th]
    \centering
    \resizebox{\textwidth}{!}{
    \begin{tabular}{c | ccccccccc}\hline
         \textbf{Method} & \textbf{MPJRE} $\downarrow$ & \textbf{MPJPE} $\downarrow$ & \textbf{MPJVE} $\downarrow$ & \textbf{Hand PE} $\downarrow$ & \textbf{Upper PE} $\downarrow$ & \textbf{Lower PE} $\downarrow$ & \textbf{Root PE} $\downarrow$ & \textbf{Jitter} $\downarrow$  & \textbf{\#FLOPs (G)} $\downarrow$ \\ \hline
         VAR-HMD \cite{Dittadi_2021_ICCV} & $4.11$ & $6.83$ & $37.99$ & -- & -- & -- & -- & -- & -- \\
        AvatarPoser \cite{jiang2022avatarposer} & $3.08$ & $4.18$ & $27.70$ & $2.12$ & $1.81$ & $7.59$ & $3.34$ & $14.49$ & $0.33$\\
        AGRoL-Diffusion \cite{du2023avatars} & $2.66$ & $3.71$ & $18.59$ & $\underline{1.31}$ & $1.55$ & $6.84$ & $3.36$ & $7.26$ & $1.00$\\
        BoDiffusion \cite{castillo2023bodiffusion} & $2.70$ & $\underline{3.63}$ & $\mathbf{14.39}$ & $\underline{1.32}$ & $\underline{1.53}$ & $7.07$ & -- & $\mathbf{4.90}$ & $0.46$\\
        EgoPoser \cite{jiang2024egoposer} & $3.09$ & $5.24$ & $24.93$ & $4.97$ & $3.79$ & $7.80$ & $3.78$ & $15.37$ & $0.33$\\ 
        MANIKIN-S \cite{jiang2024manikin} & $-$ & $\textbf{3.36}$ & $23.18$ & $\textbf{0.02}$ & $\mathbf{1.32}$ & $6.72$ & $-$ & $7.95$ &  $0.33$ \\ 
        \hline
        AGRoL-MLP (T:196) \cite{du2023avatars} & $2.69$ & $3.93$ & $22.85$ & $2.62$ & $1.89$ & $6.88$ & $3.35$ & $13.01$ & $0.88$\\
        \hline \hline 
        TW-MLP (T:41-K:2)(ours) & $\underline{2.51}$ & $\underline{3.64}$ & $19.10$ & $2.27$ & $1.73$ & $\textbf{6.40}$ & $\textbf{3.21}$ & $7.99$ & $\mathbf{0.19}$\\
        TW-MLP (T:61-K:2)(ours) & $\mathbf{2.49}$ & $3.68$ & $\underline{17.91}$ & $2.23$ & $1.77$ & $\underline{6.44}$ & $\underline{3.28}$ & $\underline{7.15}$ & $\underline{0.30}$\\
        \hline
    \end{tabular}
    }
    \caption{Comparison with real-time state-of-the-art methods (i.e., whose GFLOPs is less than $1.00$) for 3D full-body motion generation. Results are reported on AMASS dataset for different metrics. The best results are in bold and the second best results are underlined.}
    \vspace{-0.8em}
    \label{tab:sota_s1}
\end{table*}

\begin{table}[t!]
\centering
\resizebox{\columnwidth}{!}{%
\begin{tabular}{lcccc}
\hline
\textbf{Method} & \textbf{MPJRE} \darr & \textbf{MPJPE} \darr & \textbf{MPJVE} \darr & \textbf{Jitter} \darr \\
\hline
TW-MLP (T:21-K:1) & $2.60$ & $3.81$ & $22.20$ & $10.17$ \\
TW-MLP (T:21-K:2) & $2.56$ & $3.78$ & $22.19$ & $10.80$ \\
TW-MLP (T:21-K:3) & $2.58$ & $3.81$ & $22.17$ & $10.18$ \\

\hline 
TW-MLP (T:41-K:1) & $2.54$ & $3.71$ & $19.27$ & $8.06$ \\
TW-MLP (T:41-K:2) & $2.51$ & $3.64$ & $19.10$ & $7.99$ \\
TW-MLP (T:41-K:3) & $2.53$ & $3.72$ & $19.17$ & $7.98$ \\

\hline 
TW-MLP (T:61-K:1) & $2.50$ & $3.71$ & $17.97$ & $7.31$ \\
TW-MLP (T:61-K:2) & $2.49$ & $3.68$ & $17.91$ & $7.15$ \\
TW-MLP (T:61-K:3) & $2.49$ & $3.67$ & $17.82$ & $6.96$ \\
\hline \hline

AGRoL-MLP (T:196-K:0) \cite{du2023avatars} & $2.69$ & $3.93$ & $22.85$ & $13.01$ \\
AGRoL-MLP (T:61-K:0) \cite{du2023avatars} & $2.66$ & $3.99$ & $25.65$ & $13.61$ \\
AGRoL-MLP (T:41-K:0) \cite{du2023avatars} & $2.63$ & $3.85$ & $25.15$ & $17.05$ \\
AGRoL-MLP (T:21-K:0) \cite{du2023avatars} & $2.79$ & $4.20$ & $29.04$ & $19.40$ \\
        
\hline

\end{tabular}}
\caption{Impact of window lenght $T$ and the number of temporal windows $K$. For comparison, we also add the baseline results (i.e., AGRoL-MLP) where no temporal windowing is applied ($K=0$). Results are reported for four metrics.}
\label{table:ablation}
\vspace{-0.8em}
\end{table}

\subsubsection{Implementation Details}

We train our model for $300$K steps using the AdamW optimizer, with an initial learning rate of $3e-4$. The learning rate is then reduced to $1e-5$ after $225$K steps. Furthermore, a batch size of $128$ is used for training. The number of MLP blocks, denoted as $L$, is set to $10$, which balances the trade-off between accuracy and computational efficiency. Additionally, the outputs from the temporal window blocks are concatenated with the outputs from the odd-numbered MLP-blocks, allowing the model to effectively integrate past information into the representations from different parts of the MLP model.

\section{Experiments}

\subsection{Dataset}

We use the AMASS dataset~\cite{mahmood2019amass} as our benchmark for training and evaluation, enabling a comparative analysis of our method with existing state-of-the-art methods. In particular, we follow the protocol proposed by~\cite{jiang2022avatarposer} and use the CMU~\cite{CMUDataset}, BMLrub~\cite{troje2002decomposing}, and HDM05~\cite{muller2007mocap} subsets of the AMASS dataset. We split each subset into training and testing sets, allocating $90\%$ of the data for training and the remaining $10\%$ for testing, allowing for a standardized and comparable assessment of the performance of all methods.

\begin{table*}[!th]
    \centering
    \resizebox{\textwidth}{!}{%
    \begin{tabular}{lcccccccccc}
        \hline
        \textbf{Method} & $L$ & \textbf{Concat Layers} & \textbf{MPJRE} \darr & \textbf{MPJPE} \darr & \textbf{MPJVE} \darr & \textbf{Hand PE} \darr & \textbf{Upper PE} \darr & \textbf{Lower PE} \darr & \textbf{Root PE} \darr & \textbf{Jitter} \darr \\
        \hline
        AGRoL-MLP (T:196) \cite{du2023avatars}  & $12$ & - & $2.69$ & $3.93$ & $22.85$ & $2.62$ & $1.89$ & $6.88$ & $3.35$ & $13.01$ \\
        AGRoL-MLP (T:41) \cite{du2023avatars}  & $10$ & - & $2.63$ & $3.85$ & $25.15$ & $2.39$ & $1.81$ & $6.80$ & $3.43$ & $17.05$ \\
        \hline \hline
        TW-MLP (T:41-K:2)(ours) & $12$ & $1, 3, 4, 7, 9, 11$ & $2.51$ & $3.69$ &$ 19.08$  & $2.23$ & $1.73$ & $6.52$ & $3.22$ & $8.10$  \\
        TW-MLP (T:41-K:2)(ours) & $10$ & $2, 4, 6, 8, 10$ & $2.54$ & $3.74$ &$ 19.53$  & $2.36$ & $1.80$ & $6.54$ & $3.34$ & $8.41$  \\
        TW-MLP (T:41-K:2)(ours) & $10$ & $1, 3, 5, 7, 9$ & $2.51$ & $3.64$ &$ 19.10$  & $2.27$ & $1.73$ & $6.40$ & $3.21$ & $7.99 $  \\
        \hline
        
        \hline
    \end{tabular}}
    \caption{Impact of the number of MLP-Blocks $L$ and importance of concatenation layers for temporal windows. Results are reported for different metrics.}
    \vspace{-0.4em}
    \label{table:blocksize}
\end{table*}

\begin{table*}[!th]
    \centering
    \resizebox{\textwidth}{!}{
    \begin{tabular}{c | ccccccccc}\hline
         \textbf{Method} & \textbf{MPJRE} $\downarrow$ & \textbf{MPJPE} $\downarrow$ & \textbf{MPJVE} $\downarrow$ & \textbf{Hand PE} $\downarrow$ & \textbf{Upper PE} $\downarrow$ & \textbf{Lower PE} $\downarrow$ & \textbf{Root PE} $\downarrow$ & \textbf{Jitter} $\downarrow$  & \textbf{\#FLOPs (G)} $\downarrow$ \\ \hline
        AvatarJLM \cite{zheng2023realistic} & $2.90$ & $3.35$ & $20.79$ & $1.24$ & $1.42$ & $6.14$ & $2.94$ & $8.39$ & $4.64$\\
        SAGE Net \cite{feng2024stratified} & $2.53$ & $3.28$ & $20.62$ & $1.18$ & $1.39$ & $6.01$ & $2.95$ & $6.55$ & $4.10$\\
        MANIKIN-LN \cite{jiang2024manikin} & $-$ & $2.73$ & $13.55$ & $0.01$ & $1.30$ & $5.13$ & $-$ & $7.95$ &$4.64$\\ 
        MMD \cite{dong2024realistic} & $2.30$ & $3.17$ & $17.32$ & $0.79$ & $1.25$ & $5.94$ & $2.86$ & $6.52$ &$7.98$\\ 

        \hline \hline 
        TW-MLP (T:41-K:2)(ours) & $2.51$ & $3.64$ & $19.10$ & $2.27$ & $1.73$ & $6.40$ & $3.21$ & $7.99$ & $0.19$\\
        TW-MLP (T:61-K:2)(ours) & $2.49$ & $3.68$ & $17.91$ & $2.23$ & $1.77$ & $6.44$ & $3.28$ & $7.15$ & $0.30$\\
        \hline
    \end{tabular}
    }
    \caption{Comparison with complex state-of-the-art methods (i.e., whose GFLOPs is higher than $1.00$) for 3D full-body motion generation. Results are reported on AMASS dataset for different metrics. As noticed, the complex of our method is improved by $\times 20$ in term of GFLOPs compared to complex (non-real-time) baseline methods while obtaining compatible reconstruction performance. }
    \vspace{-0.8em}
    \label{tab:sota_s2}
\end{table*}

\begin{table}[t!]
    \centering
    \resizebox{\columnwidth}{!}{%
    \begin{tabular}{c|c c c}
        \hline
        \textbf{Method} & \textbf{FLOPs \darr} & \textbf{SIZE (MB) \darr} & \textbf{\#PARAMS (M) \darr} \\ \hline
        AvatarPoser \cite{jiang2022avatarposer}& $0.33$G & $15.73$ & $4.12$ \\ 
        AGRoL-Diffusion \cite{du2023avatars} & $1.00$G & $28.54$ & $7.48$ \\
        AvatarJLM \cite{zheng2023realistic} & $4.64$G & $243.41$ & $63.81$ \\
        BoDiffusion \cite{castillo2023bodiffusion} & $0.46$G & $83.70$ & $21.94$ \\ 
        SAGE-Net \cite{feng2024stratified} & $4.10$G & $458.79$ & $120.27$\\ 
        EgoPoser \cite{jiang2024egoposer} & $0.33$G & $15.77$ & $4.12$\\
        MMD \cite{dong2024realistic} & $7.98$G & $853.47$  & $101.72$\\
        MANIKIN-S \cite{jiang2024manikin} & $0.33$G & $-$ & $4.12$ \\
        MANIKIN-L \cite{jiang2024manikin} & $4.64$G & $-$ & $63.8$ \\
        MANIKIN-LN \cite{jiang2024manikin} & $4.64$G & $-$ & $63.8$ \\
        \hline
        AGRoL-MLP (T:196) \cite{du2023avatars} & $0.88$G & $14.25$ & $3.74$ \\
        \hline \hline
        TW-MLP (T:41-K:2) (ours) & $\mathbf{0.18}$G & $\mathbf{12.10}$ & $\mathbf{3.17}$ \\ 
        TW-MLP (T:61-K:2) (ours) & $0.30$G & $12.47$ & $3.27$ \\ 
        \hline
    \end{tabular}}
    
    \caption{Model size and model complexity. We report FLOPs, model sizes and number of parameters for different methods.}
    \label{tab:size_flops_params}
    \vspace{-0.4em}
\end{table}

\begin{table}[t!]
    \centering
    \resizebox{\columnwidth}{!}{%
    \begin{tabular}{c|c c}
        \hline
        \textbf{Method} & \textbf{CPU (ms) \darr} & \textbf{CPU (FPS) \uparr} \\ \hline
        AvatarPoser~\cite{jiang2022avatarposer} & $6.0$ & $\mathbf{72.0}$ \\ 
        AGRoL-MLP (T:196)~\cite{du2023avatars} & $30.1$ & $26.0$ \\
        AGRoL-Diffusion~\cite{du2023avatars} & $290.5$ & $3.5$ \\\hline
        TW-MLP (T:41-K:2) (ours) & $\mathbf{5.9}$ & $\mathbf{72.0}$ \\
        TW-MLP (T:61-K:2) (ours) & $6.6$ & $72.0$ \\
        \hline
    \end{tabular}}

    \caption{On-device inference time. Results are reported on the Meta Quest-3 heads on latency (ms) and FPS for different methods.}
    \label{tab:placeholder_label}
    \vspace{-0.8em}
\end{table}

 \subsection{Metrics}

To evaluate the performance of all methods, we employ a set of standard metrics commonly used in the literature~\cite{jiang2022avatarposer, du2023avatars}. These metrics can be categorized into four groups:

\noindent
\textbf{Rotation-related metric:} We use the Mean Per Joint Rotation Error (MPJRE) to measure the average relative rotation error across all joints, expressed in degrees. This metric provides insight into the accuracy of methods in predicting joint rotations.

\noindent
\textbf{Velocity-related metrics:} We utilize two metrics to evaluate the velocity aspects of the performance of methods. The Mean Per Joint Velocity Error (MPJVE) calculates the average velocity error across all joint positions, measured in cm/s. Additionally, we use Jitter to assess the smoothness of the motion by computing the rate of change in acceleration over time for all body joints in the global space, with units of $(10^2m/s^3)$.

\noindent
\textbf{Position-related metrics}: We employ a range of metrics to evaluate the positional accuracy of methods. The Mean Per Joint Position Error measures the average variation in position across all joint locations. Furthermore, we use specific metrics to assess the positional error of different body parts, including Root PE (root joint), Hand PE (average error for both hands), Upper PE (joints in the upper body), and Lower PE (joints in the lower body). These metrics provide a comprehensive understanding of the performance of methods in predicting joint positions.

\noindent
\textbf{Computation-related metrics}: Floating-Point Operations Per Second (FLOPs) is a metric used to measure the computational performance of models. It represents the number of million floating-point operations, serving as an indicator of the model processing speed.

\subsection{Comparison with State-of-the-Art}

First, we compare our method with existing real-time state-of-the-art methods, which are built upon various NN architectures and GFLOPs is less than $1.00$. Tab.~\ref{tab:sota_s1} presents the performance of these baselines, including our method (i.e., TW-MLP), across multiple evaluation metrics. We also report two variants of our method, with window length of $T=41$ and $T=61$. Note that the base method AGRoL-MLP~\cite{du2023avatars}, which uses a longer sequence length ($T=196$) with a MLP model, is included in the table for comparison.

Notably, our method outperforms the base MLP-method, AGRoL-MLP, in terms of both accuracy and complexity. Specifically, our method yields better results on the Jitter metric, which indicates smoother motion generation, and improves the metric score from $13.01$ to $7.99$ for the $T=41$ configuration. Furthermore, our method not only improves the generation accuracy, but it also provides a benefit in computational cost. In terms of FLOPs, the computational cost is substantially decreased from 0.88 to 0.19 for our method.

\subsection{Ablation Study}

In ablation study, we  conduct additional experiments  to understand how some parameters can impact the performance of our method. Additionally, we evaluate and compare our method to other baseline methods in terms of model size, complexity, and on-device inference time.

\subsubsection{Impact of $T$ and $K$ Parameters}

In this section, we present an ablation study to analyze how window length $T$ and the number of temporal windows $K$ can influence our results. For this purpose, we train multiple models under different configurations for $T \in\{21, 41, 61\}$ and $K \in \{1, 2, 3\}$ values. The results for this experiment are reported in Tab.~\ref{table:ablation}. From the results, $T=41,K=2$ and $T=61,K=2$ configurations yield the best performance compared to others. Furthermore, the accuracy of our model for different $K$ starts to saturate after $K=3$ for almost all $T$ configurations. This indicates that for longer temporal windows, the past information starts to be distracting for the generation step. On the other hand, increasing the $T$ value has a positive impact on performance. However, while larger $T$ values improve rotation and velocity-related metrics, they can actually lead to decreased performance in position-related metrics as summarized in Tab.~\ref{tab:sota_s1}. Intuitively, this indicates that incorporating too much past context can be detrimental, causing our method to lose its generalization capability and perform poorly on other metrics.

Furthermore, we report the results for AGRoL-MLP baseline method with varying window lengths. Notably, our windowing mechanism provides substantial improvements in accuracy using similar window lengths compared to the baseline model.   

\subsubsection{Impact of $L$ and Concatenation Layers}

In this section, we analyze the impact of numbers of MLP-blocks $L$ and the importance of concatenation layers for temporal windows. We conduct this experiment using the configuration $T=41,K=2$. The results are summarized in Tab.~\ref{table:blocksize}. The results indicate that the concatenation layers for temporal windows have an impact on the performance for our method. This is because progressively incorporating past context information into the calculation of latent representations leads to more effective learning. Additionally, increasing the number of the MLP-blocks results in a slight decrease in performance, indicating that larger blocks may not necessarily lead to better performance.

\begin{figure*}[!htb]
\centering
\begin{tabular}{ccccc}

\includegraphics[width=0.22\textwidth,trim= 250 300 0 100,clip]{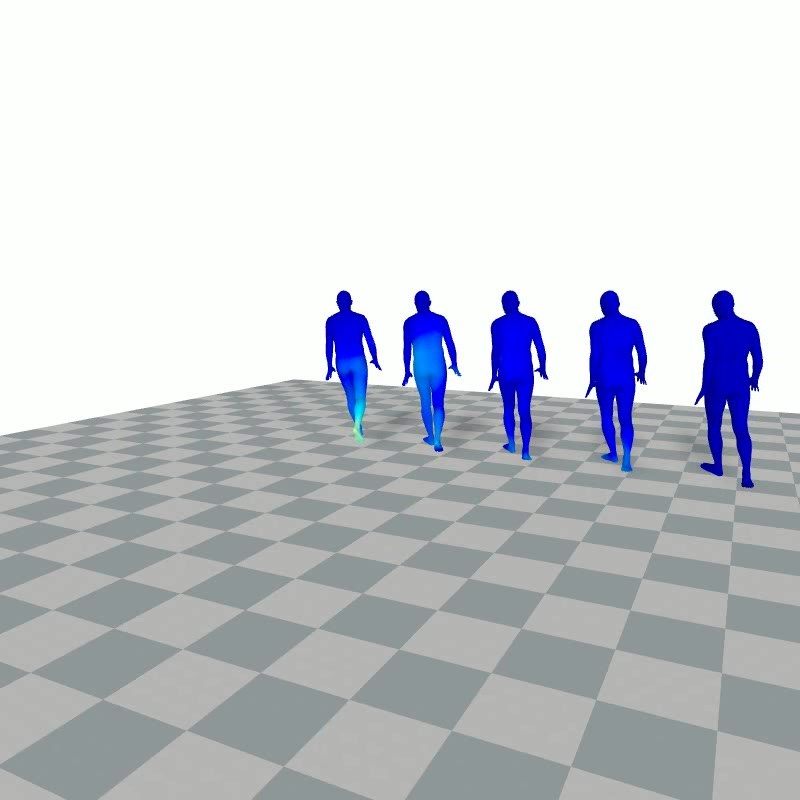} &
\includegraphics[width=0.22\textwidth, trim= 250 300 150 200,clip]{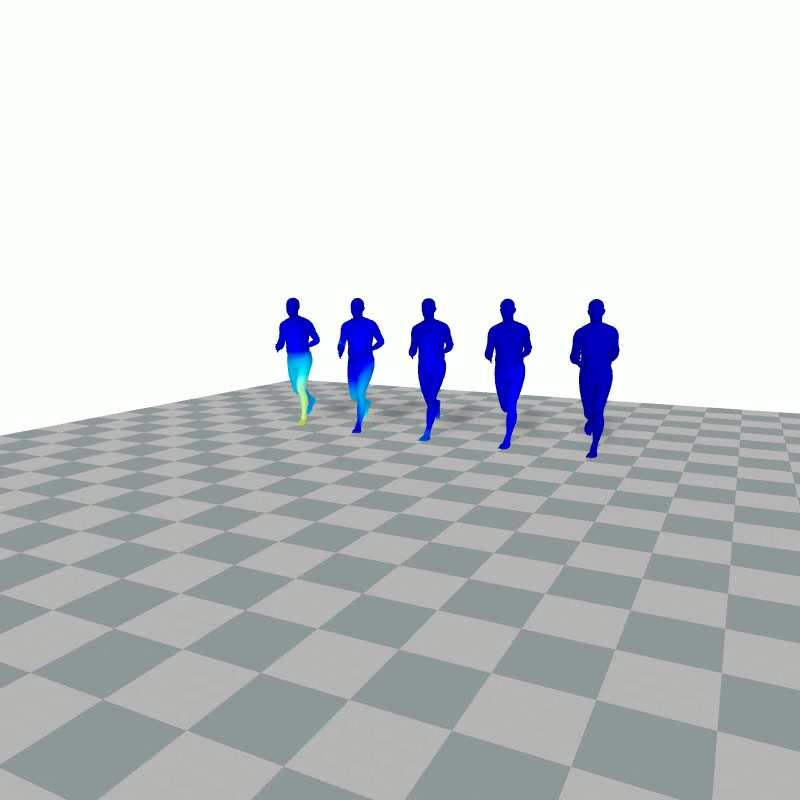} &
\includegraphics[width=0.22\textwidth, trim= 250 300 150 200,clip]{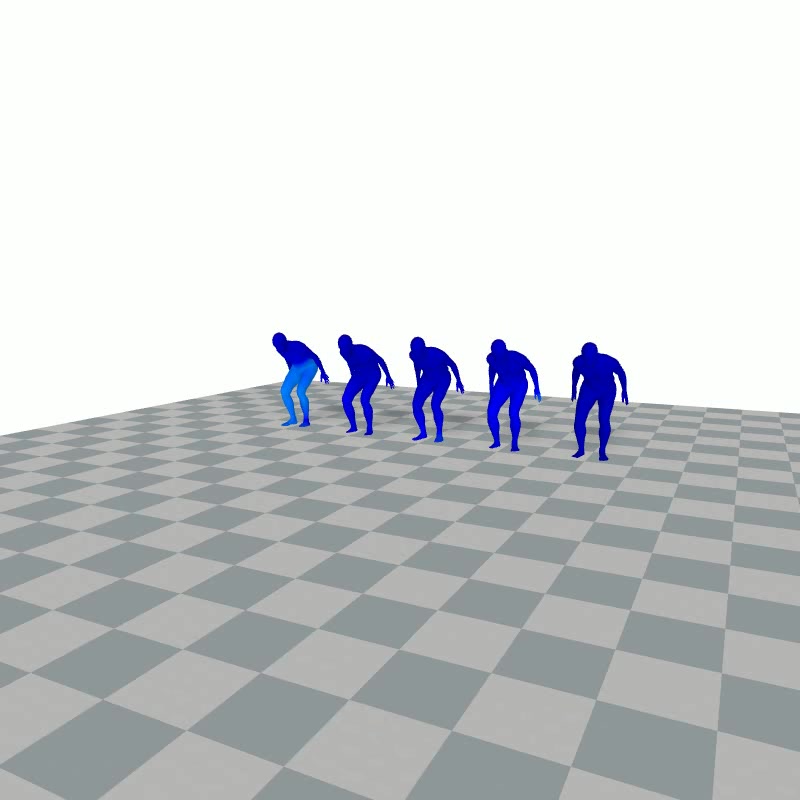} &
\includegraphics[width=0.22\textwidth, trim= 120 250 150 250,clip]{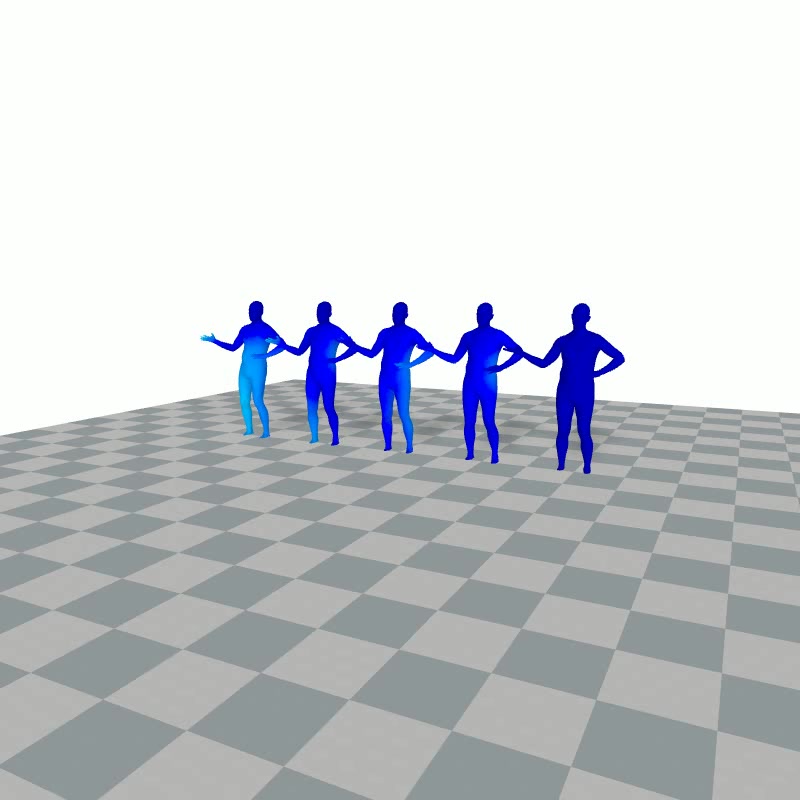} &
\includegraphics[width=0.04\textwidth, height=0.12\textheight]{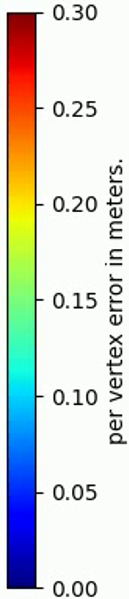} \\

\includegraphics[width=0.22\textwidth,trim= 250 320 0 100,clip]{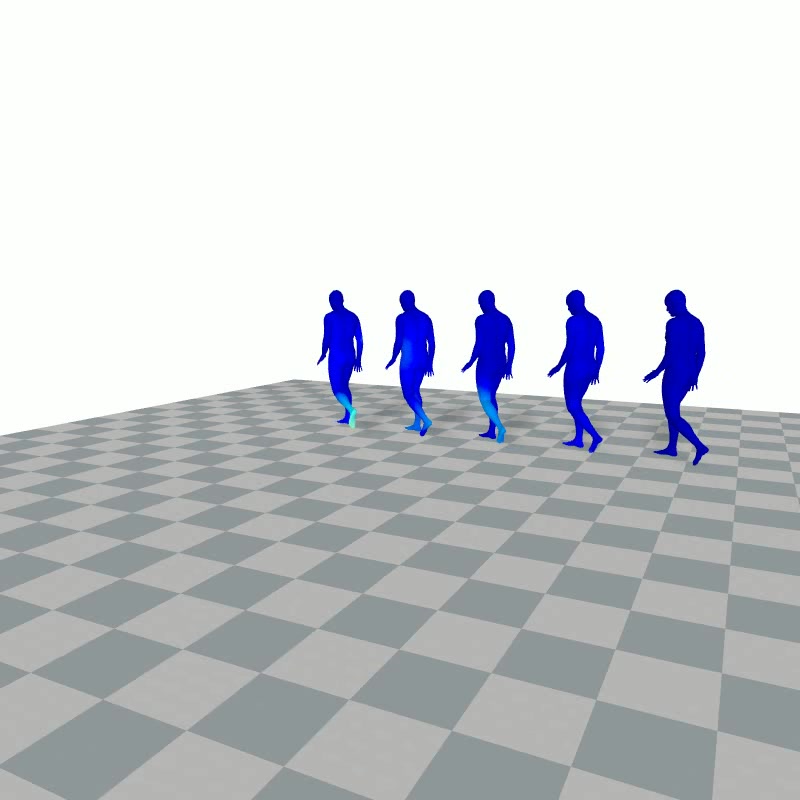} &
\includegraphics[width=0.22\textwidth, trim= 250 300 150 200,clip]{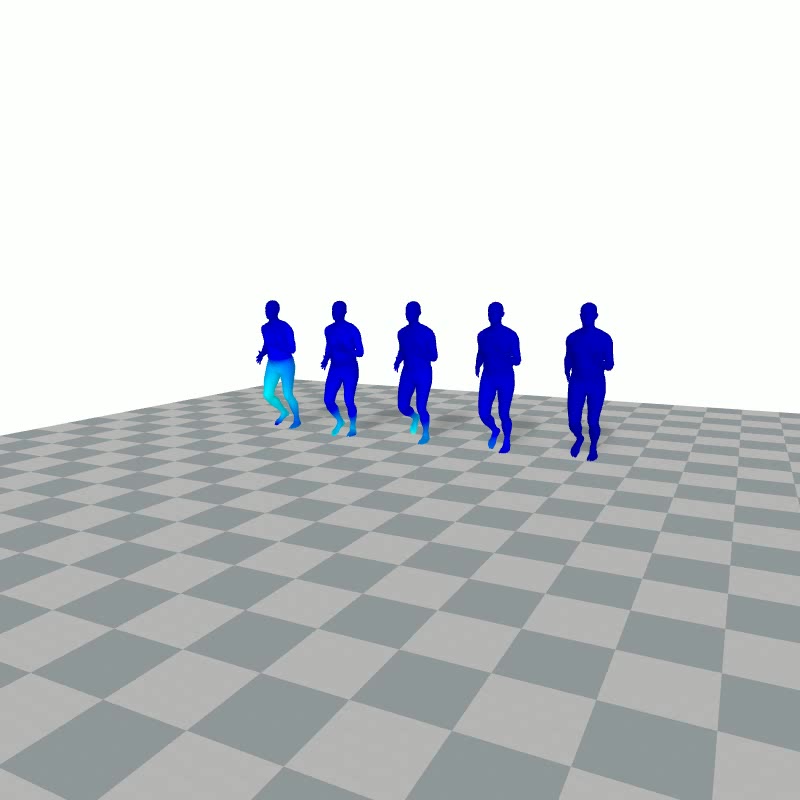} &
\includegraphics[width=0.22\textwidth, trim=250 300 150 200,clip]{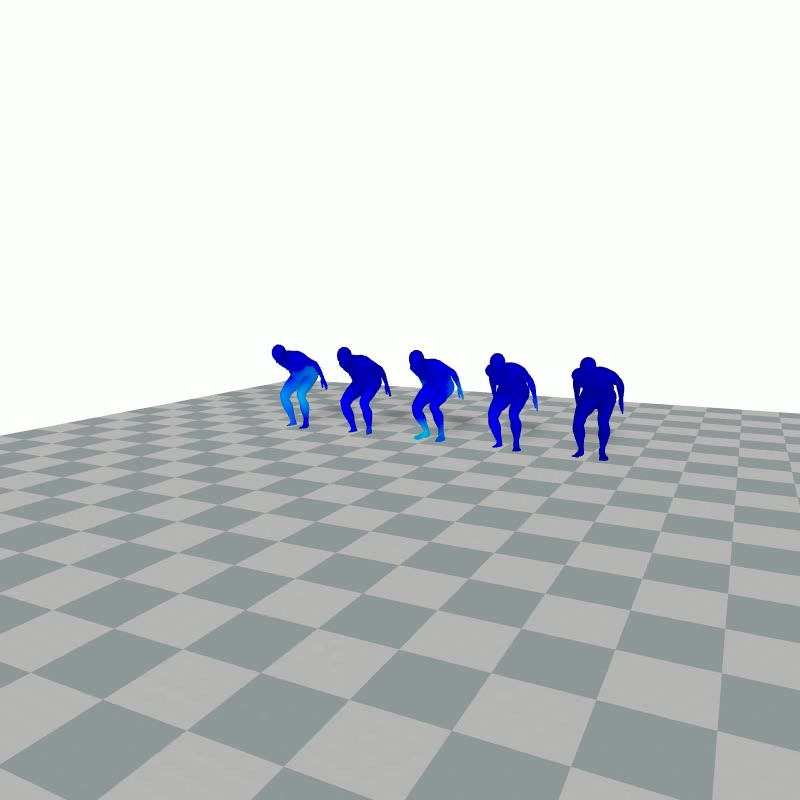} &
\includegraphics[width=0.22\textwidth, trim= 120 250 150 250,clip]{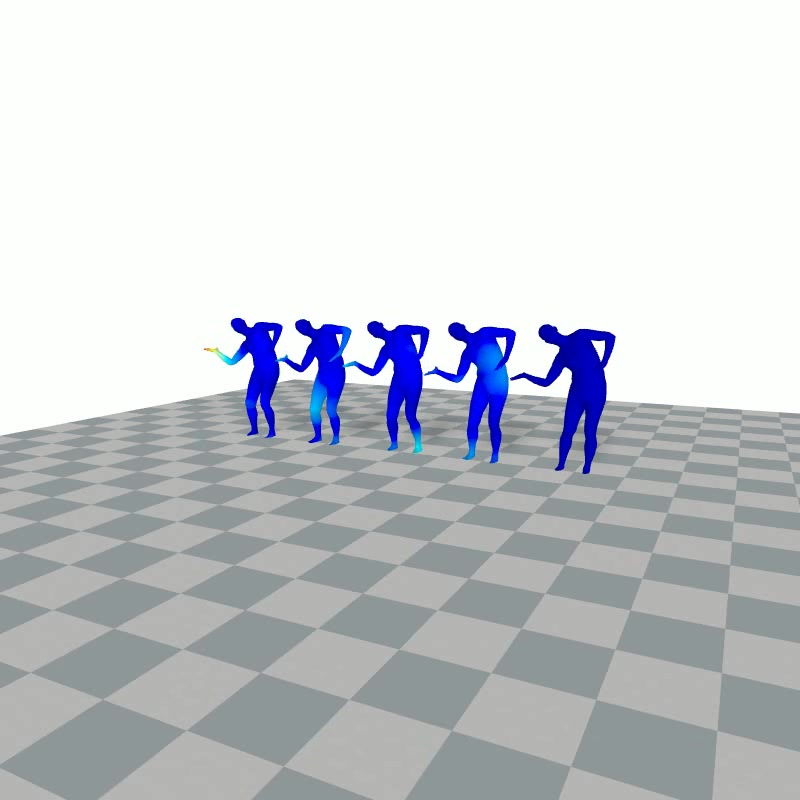} \\

\includegraphics[width=0.22\textwidth,trim= 200 320 150 100,clip]{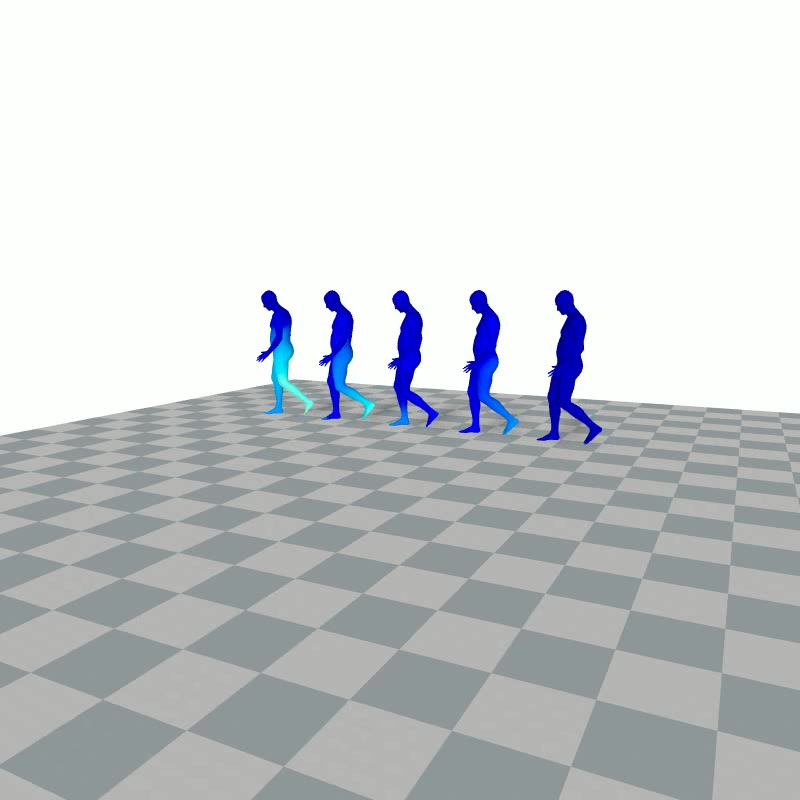} &
\includegraphics[width=0.22\textwidth, trim= 120 250 150 250,clip]{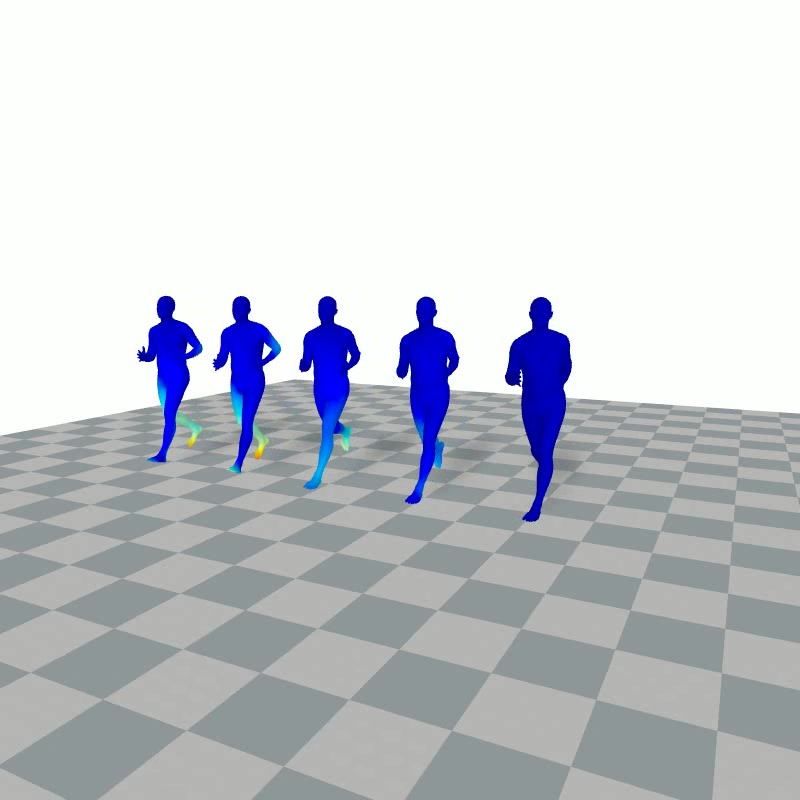} &
\includegraphics[width=0.22\textwidth, trim= 250 300 150 200,clip]{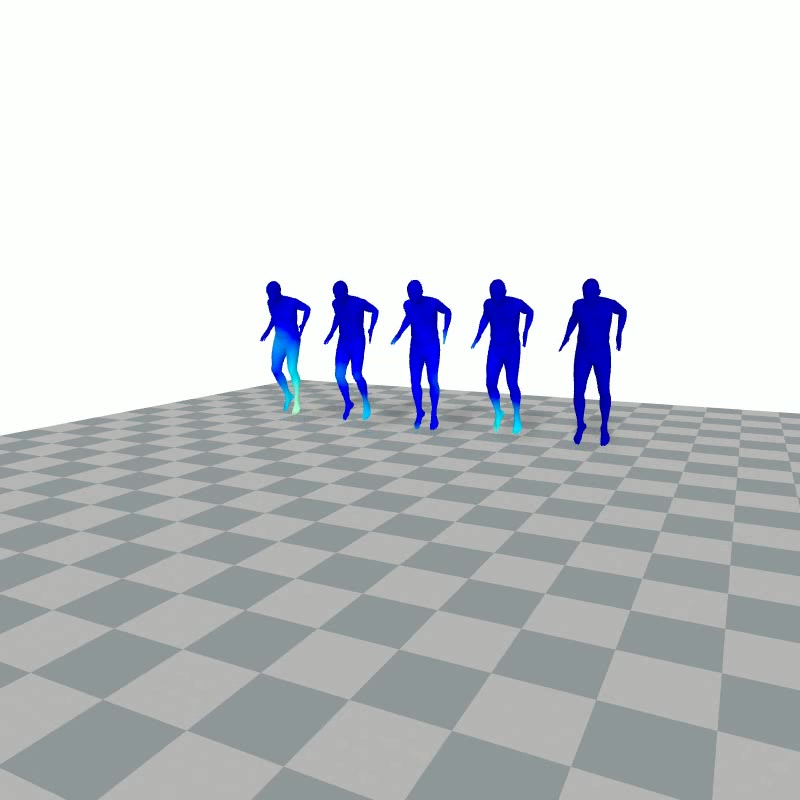} &
\includegraphics[width=0.22\textwidth, trim= 120 250 150 250,clip]{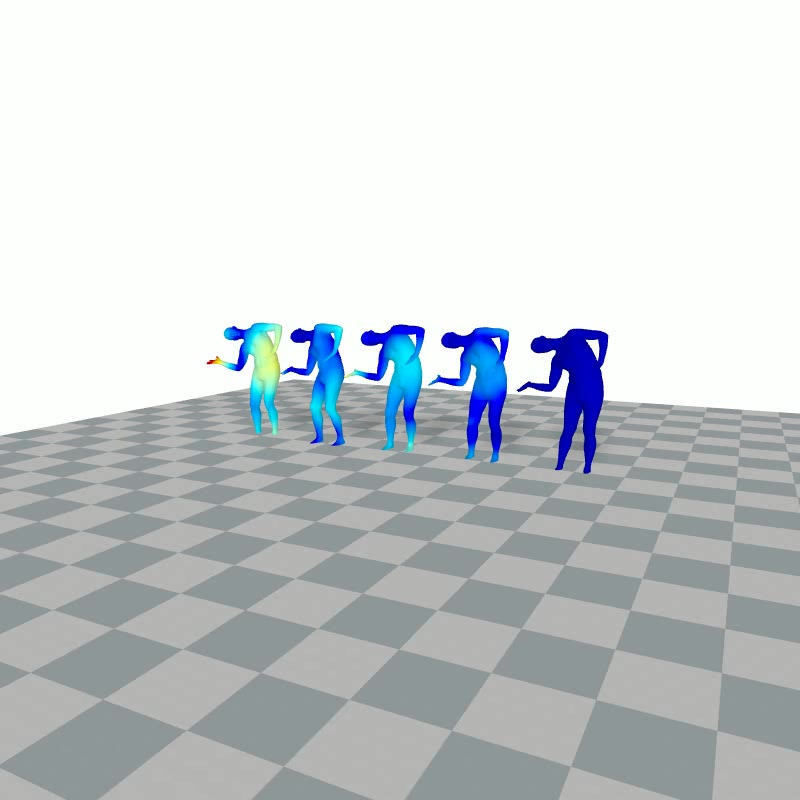} \\
\end{tabular}

\caption{Results for different motions: from left to right columns: circle-walking, fast-runing, jumping-on-the-stop and side-bend. Heat map error results are illustrated for AGRoL-MLP (T:196), AGRoL-Diffusion \cite{du2023avatars}, TW-MLP (T:41-K:2), SAGE-Net \cite{feng2024stratified} and the ground-truth.}
\vspace{-0.8em}
\label{Fig:result_poses_1}
\end{figure*}

\begin{figure*}[!htb]
\centering
\begin{tabular}{ccccc}

\includegraphics[width=0.22\textwidth, trim= 150 300 150 250,clip]{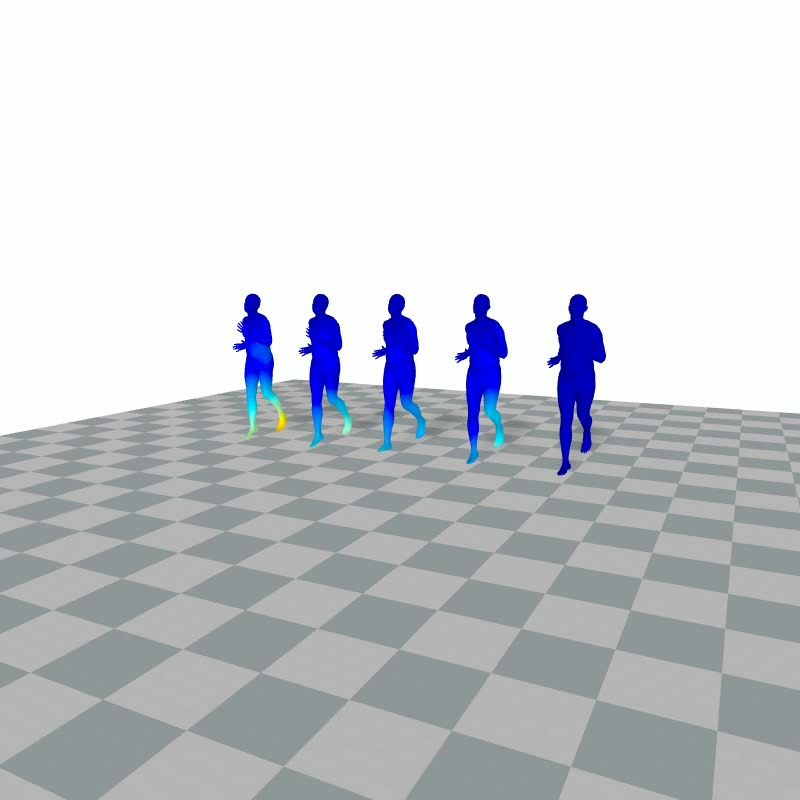} &
\includegraphics[width=0.22\textwidth,trim= 150 300 200 300,clip]{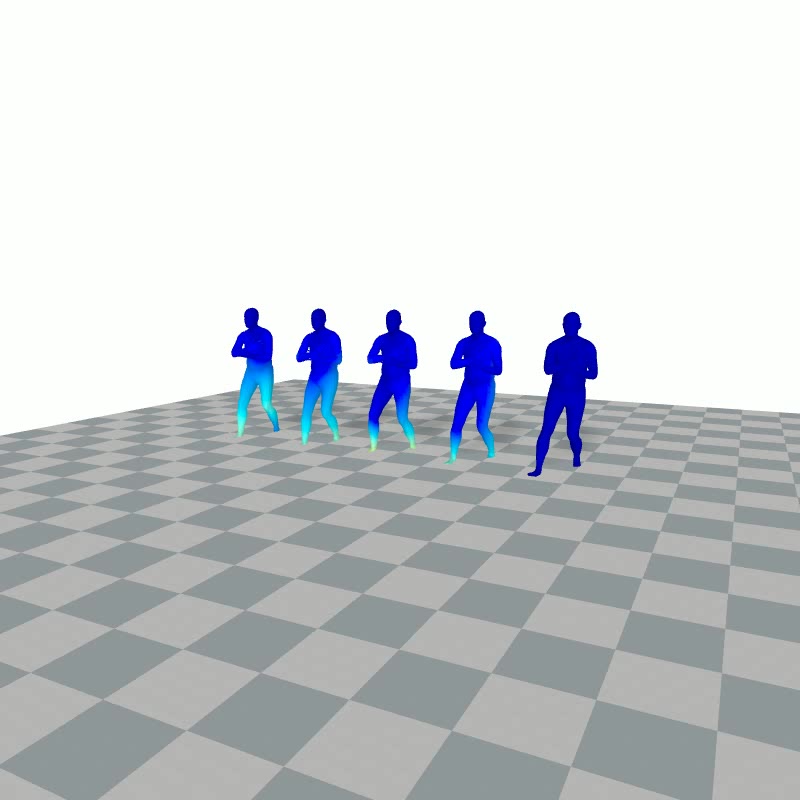} &
\includegraphics[width=0.22\textwidth, trim= 200 300 150 200,clip]{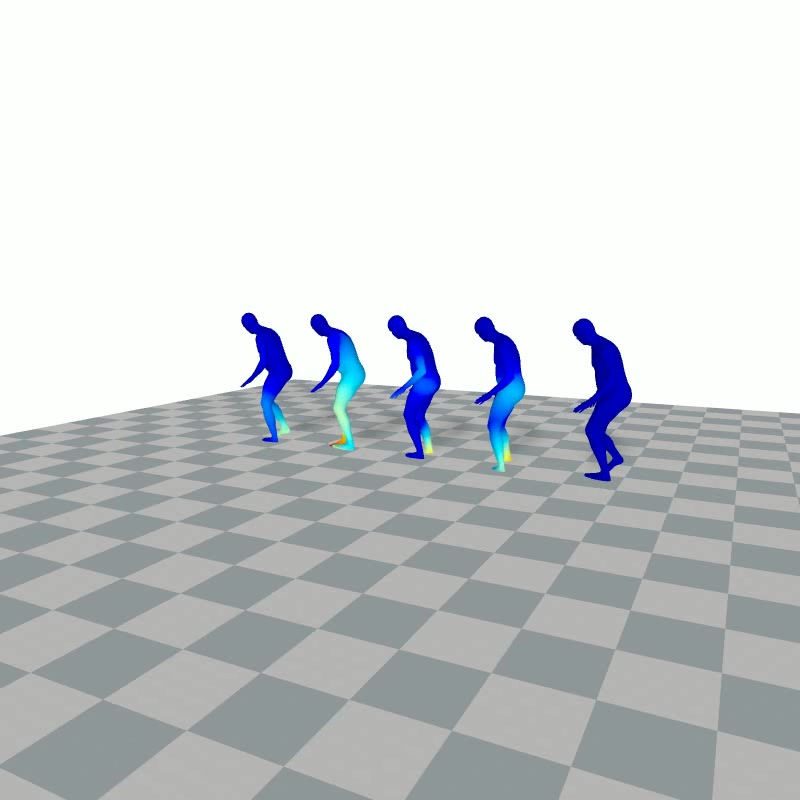} &
\includegraphics[width=0.22\textwidth, trim= 100 300 150 200,clip]{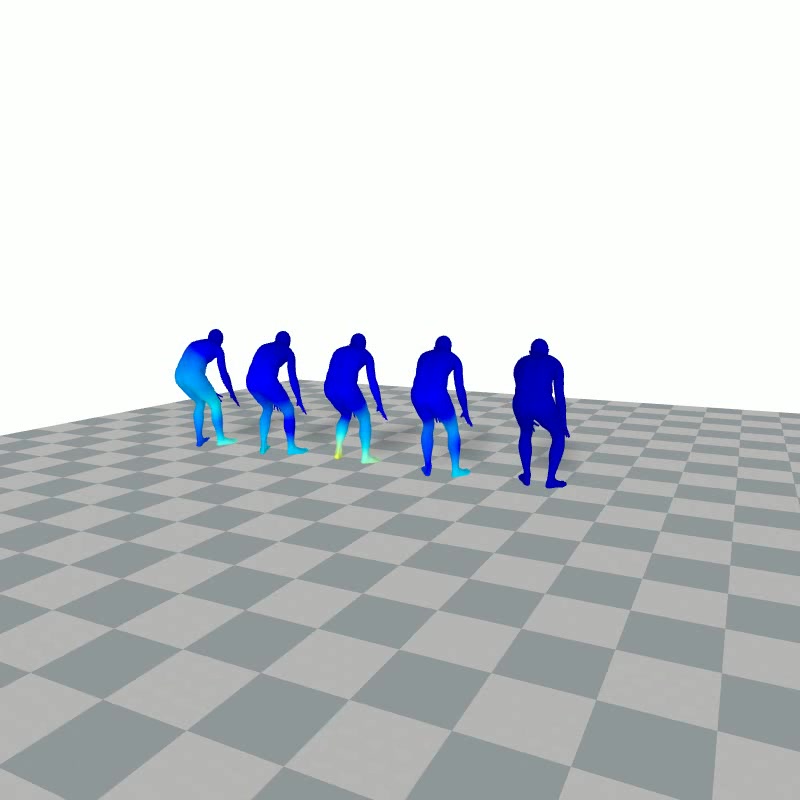} & 
\includegraphics[width=0.04\textwidth, height=0.12\textheight]{images/renderings/HeatMap_ErrorRange.png} \\

\includegraphics[width=0.22\textwidth, trim= 150 300 150 250,clip]{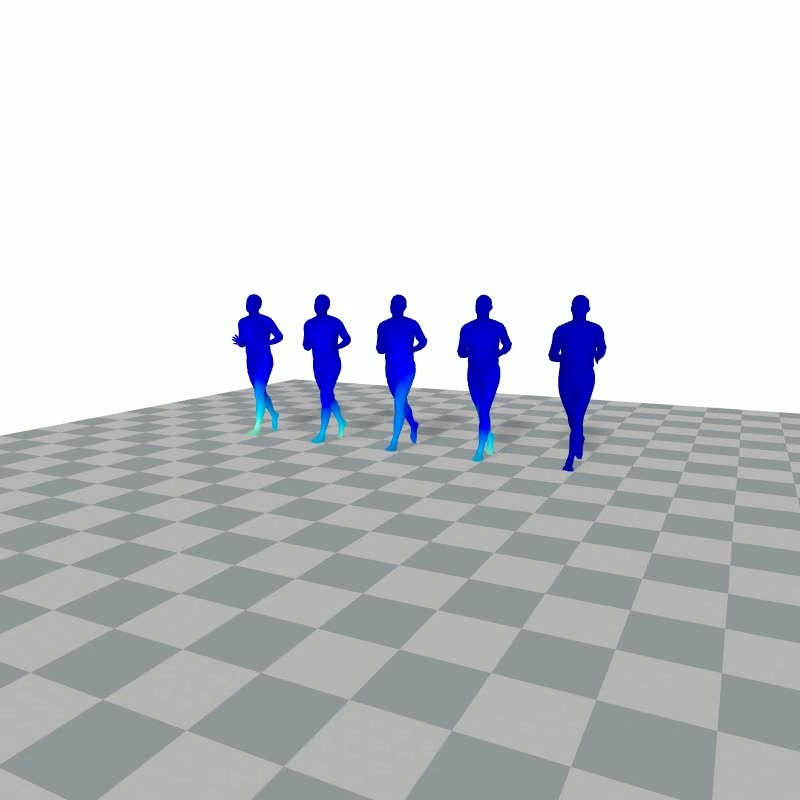} &
\includegraphics[width=0.22\textwidth,trim= 150 300 200 300,clip]{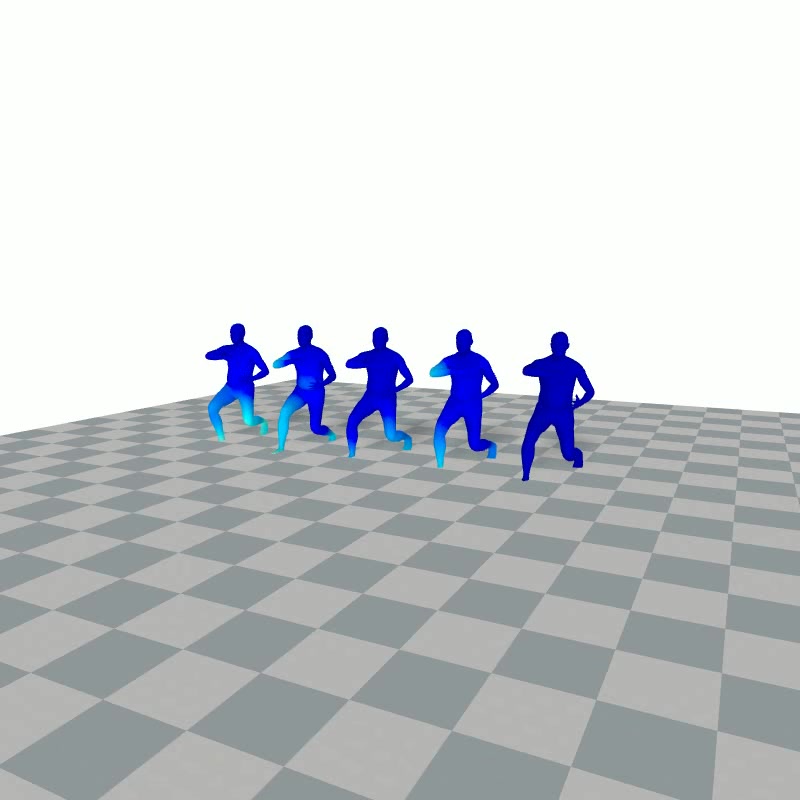} &
\includegraphics[width=0.22\textwidth, trim= 200 300 150 200,clip]{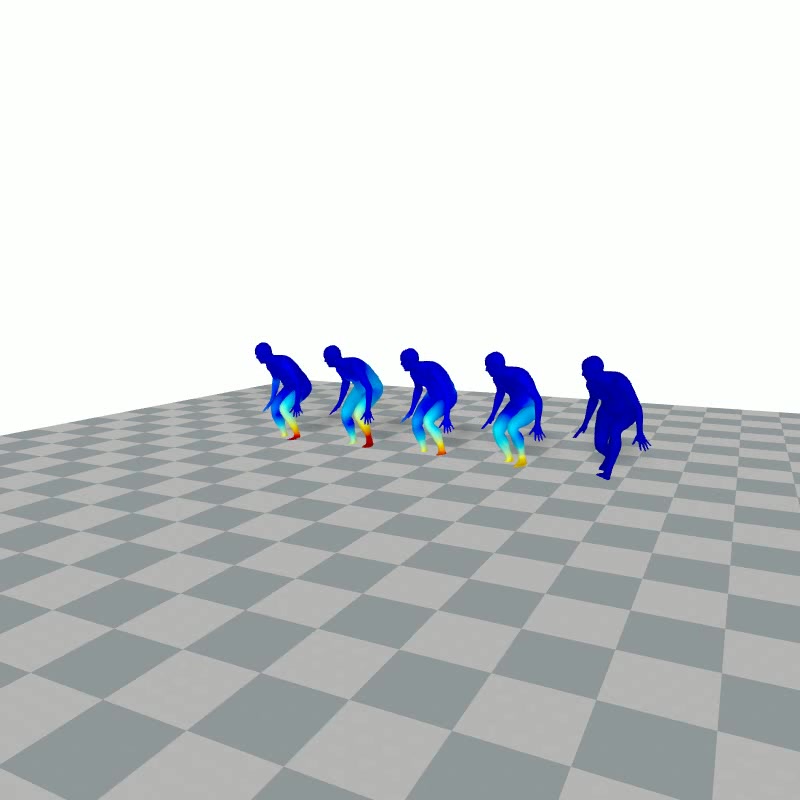} &
\includegraphics[width=0.22\textwidth, trim=100 300 150 200,clip]{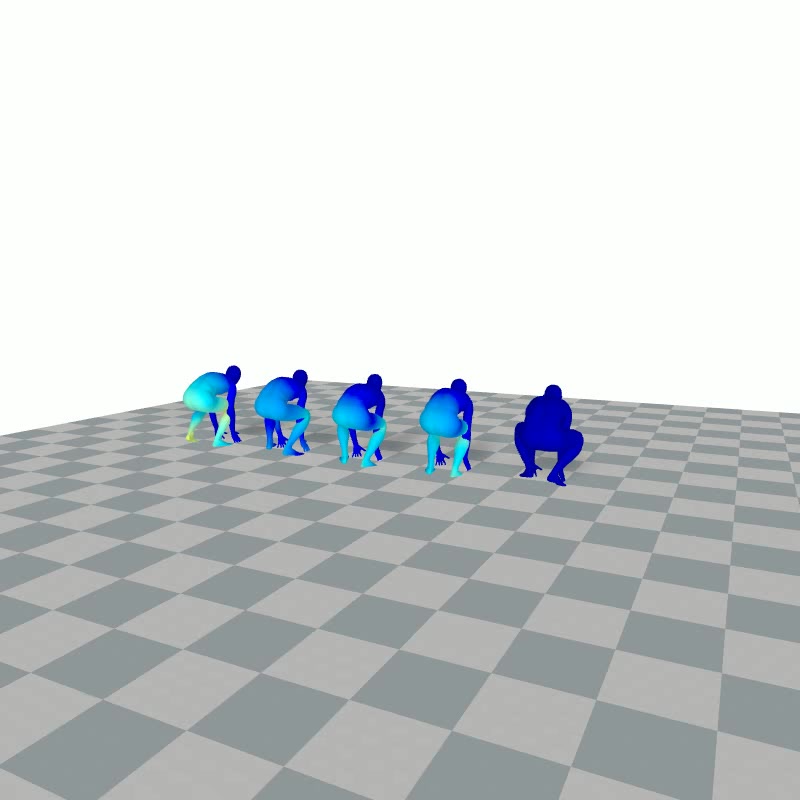} \\

\includegraphics[width=0.22\textwidth, trim= 150 300 150 250,clip]{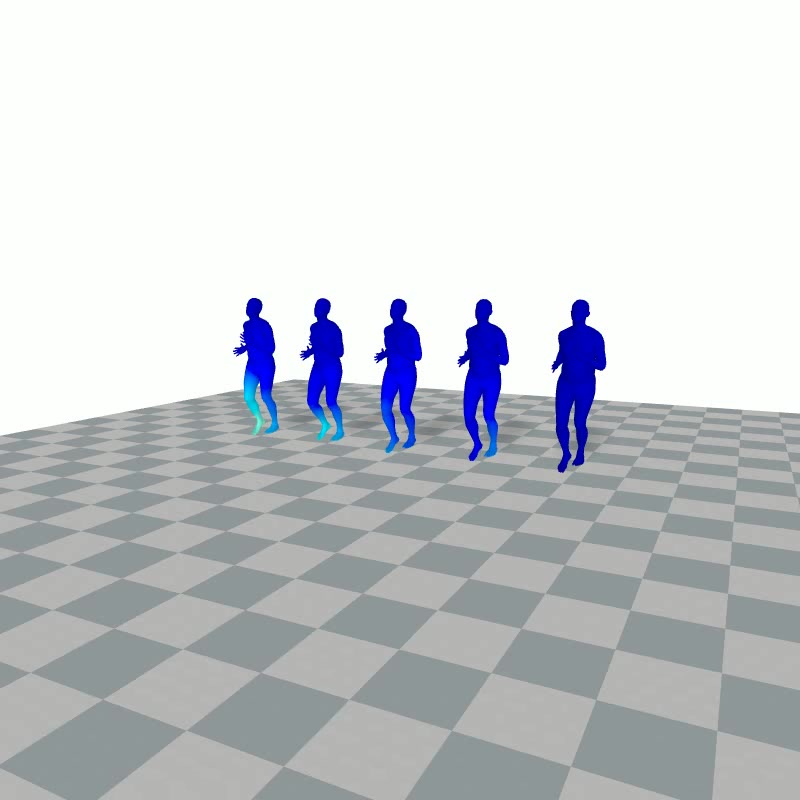} &
\includegraphics[width=0.22\textwidth,trim= 150 300 200 300,clip]{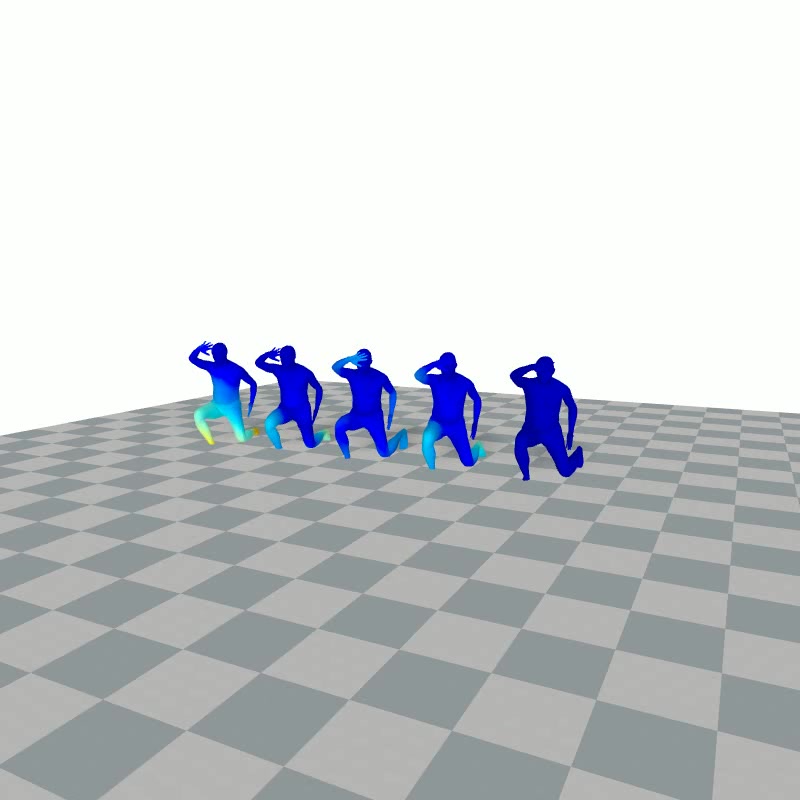} &
\includegraphics[width=0.22\textwidth, trim= 200 300 120 200,clip]{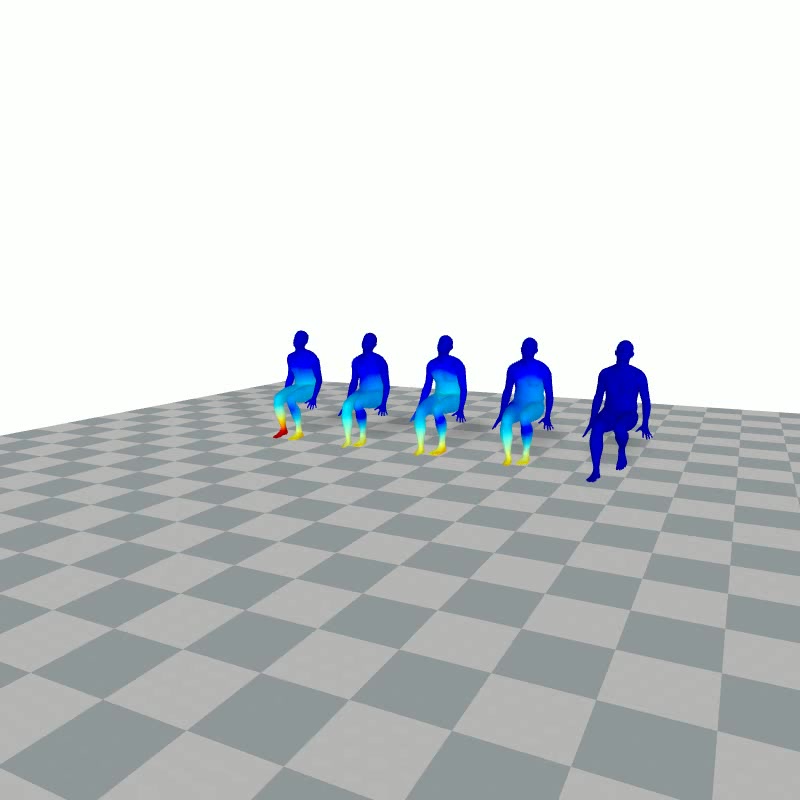} &
\includegraphics[width=0.22\textwidth, trim=100 300 150 200,clip]{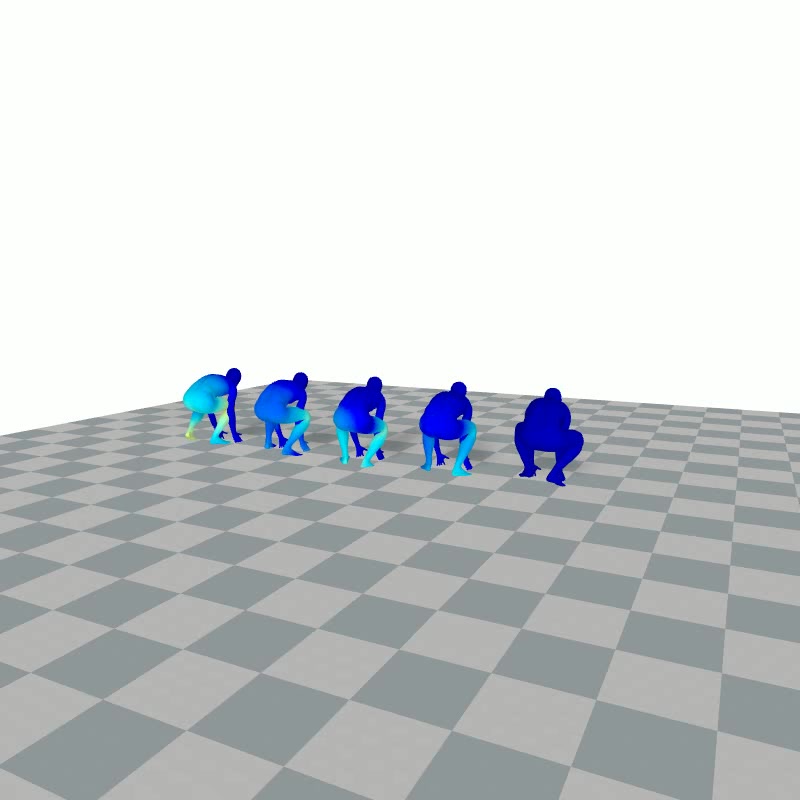} \\

\end{tabular}

\caption{Results for different motions: from left to right columns: slow-running, stretching, sitting and lifting. Heat map error results are illustrated for AGRoL-MLP (T:196), AGRoL-Diffusion \cite{du2023avatars}, TW-MLP (T:41-K:2), SAGE-Net \cite{feng2024stratified} and the ground-truth.}
\vspace{-0.8em}
\label{Fig:result_poses_2}
\end{figure*}

\subsubsection{Comparison with Complex Baselines}

In Tab.~\ref{tab:sota_s2}, we report the accuracy and computational cost for complex state-of-the-art methods. The results demonstrate that both of our configurations obtain compatible performance to highly complex methods, such as transformer and diffusion-based models. In particular, our model reduces the computational overhead by $\times 20$ in term of GLOPs compared to other complex baselines.   

\subsubsection{Model Size and Model Complexity}

Another crucial aspect of generating 3D full-body motion is the model size, complexity, and parameter size. Tab.~\ref{tab:size_flops_params} summarizes the results of these aspects for various methods, including two variants of our method ($T=41,K=2$ and $T=61,K=2$). From these results, our method obtains the best performance (i.e., low computational and memory overhead) compared to other methods.  Notably, compared to the base MLP-method, AGRoL-MLP, the FLOPs are reduced more than $3$ times for our method from $0.88$G to $0.18$G.  When we jointly interpret these results with our accuracy results, it is clear that our method can successfully balance the trade-off between accuracy and complexity.

\subsubsection{On-Device Inference Time}

Additionally, we report the inference time of various methods on the Meta Quest-3 headset by measuring latency and FPS. These results are presented in Tab.~\ref{tab:placeholder_label}. Note that all methods are first converted to ONNX models in the deployment step and executed on a single CPU. Ultimately, this experiment assesses the performance of each method in resource-constrained environments. The results demonstrate that our method runs $72$ FPS with the lowest latency on the device, outperforming other baseline methods. For the configuration $T=61,K=2$, even if latency is increased slightly, it still maintains real-time performance. 

\subsubsection{Qualitative Results}

The 3D full-body motion generation results for various methods  and different types of motion are visualized in Figs.~\ref{Fig:result_poses_1} and ~\ref{Fig:result_poses_2}. Specifically, heat maps are provided to highlight the error distribution across different body parts for each method, allowing for a detailed comparison of their performance.

\section{Conclusion}

In this paper, we propose a NN-based method that balances the trade-off between accuracy and complexity for 3D full-body motion generation. More precisely, a novel NN-mechanism is introduced that divides the longer sequences of sparse tracking inputs into smaller temporal windows and combines the current motion with past temporal windows through latent representations. In particular, our method leverages MLP-based models for efficient and effective motion generation. The experiment results on the AMASS dataset demonstrate the superiority of our method over SOTA methods, while substantially reducing computational costs and memory requirements, making our method suitable for real-world applications on edge devices.


{\small
\bibliographystyle{ieee_fullname}
\bibliography{egbib}
}

\end{document}